\documentclass[review]{elsarticle}

\usepackage{lineno,hyperref}
\usepackage{times}
\usepackage{epsfig}
\usepackage{graphicx}
\usepackage{amsmath}
\usepackage{amssymb}
\usepackage{multirow}
\usepackage{rotating}
\usepackage{booktabs}
\usepackage{float}

\usepackage{verbatim}

\usepackage{cleveref}
\crefname{section}{§}{§§}
\Crefname{section}{§}{§§}

\usepackage{amssymb}
\journal{Computer Vision and Image Understanding}

\begin{document}

\begin{frontmatter}



\title{Scene Labeling Through Knowledge-Based Rules Employing Constrained Integer Linear Programing} 

\author{ Nasim Souly, Mubarak Shah}
 \address{Center for Research in Computer Vision, University of Central Florida, USA\\ nsouly@eecs.ucf.edu, shah@crcv.ucf.edu}

\begin{abstract}
 Scene labeling task is to segment the image into meaningful regions and categorize them into classes of objects which comprised the image. Commonly used methods typically find the local features for each segment and label them using classifiers.
Afterward, labeling is smoothed in order to make sure that neighboring regions receive similar labels. However, they ignore expressive and non-local dependencies among regions due to expensive training and inference. In this paper, we propose to
use high level knowledge regarding rules in the inference to incorporate dependencies among regions in the image to improve scores of classification. Towards this aim, we extract these rules from data and transform them into constraints for Integer
Programming to optimize the structured problem of assigning labels to super-pixels (consequently pixels) of an image. In addition, we propose to use soft-constraints in some scenarios, allowing violating the constraint by imposing a penalty, to make the model more flexible. We assessed our approach on three datasets and obtained promising results.

\end{abstract}

\end{frontmatter}


\section{Introduction}
\begin{figure}[t]
\begin{center}

  \centering
  \includegraphics[width=.7\textwidth]{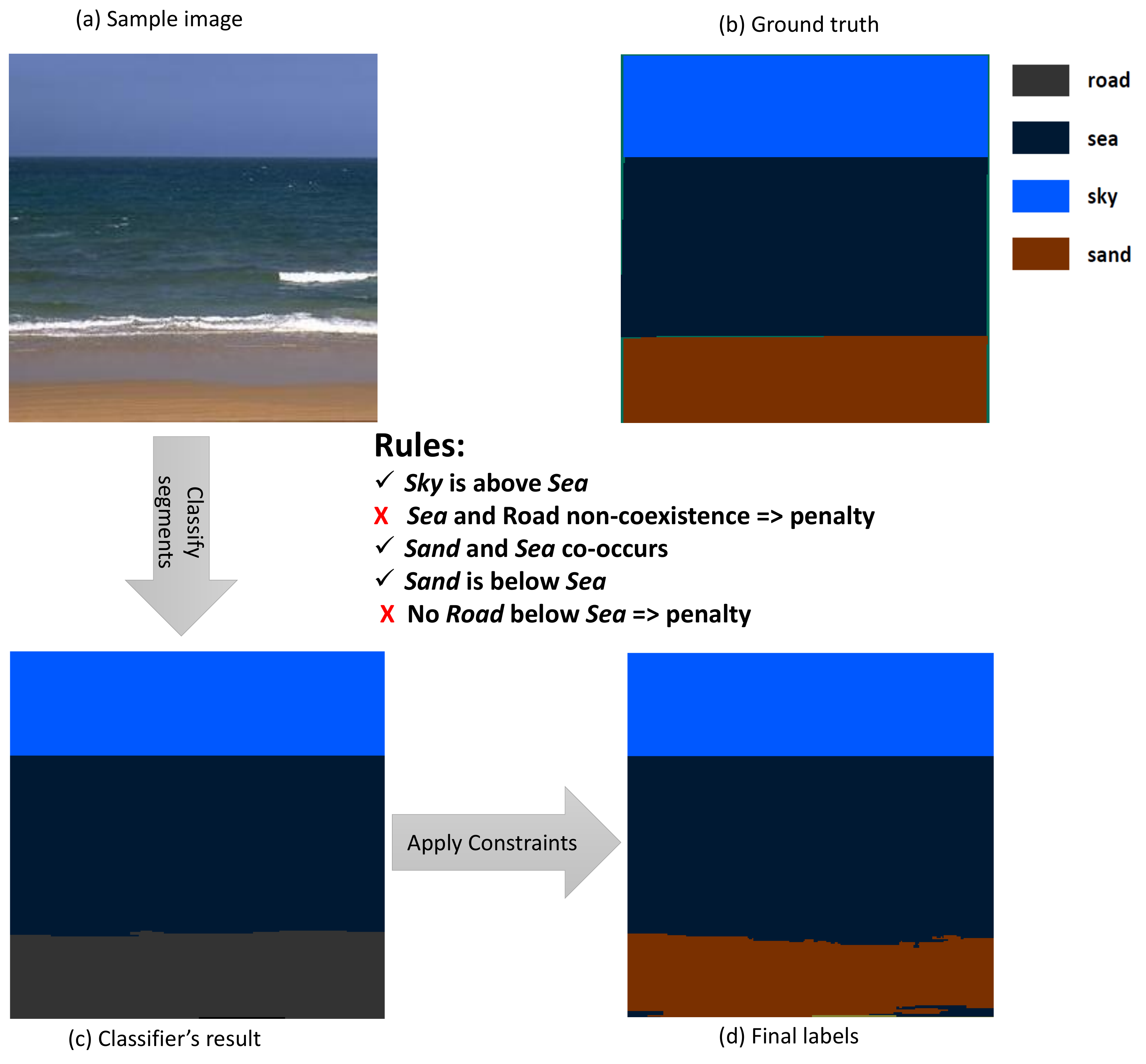}\\
  \caption{  Given an image and initial region labels obtained by the classifiers, we aim to improve  the labels using constraints learned from  the data.  (a) shows a query image, (b) shows the human annotated image (Ground Truth), (c) are labels obtained by classifiers, and (d) shows our results after enforcing constraints. }\label{intro}
    \end{center}

\end{figure}

 Semantic segmentation is one of the essential tasks to interpret and analyze an image; objects types and their placement in the scene provide the substantial information about the 3D world captured in the image. The constraints implied by the relations between different type of objects and the geometry of the objects in the scene can aid in processing and understating images. The effectiveness and advantage of using world knowledge in understanding images have been explored and assessed in early vision researches \cite{parma1981experiments}, and we want to exploit this concept to label the pixels of an image. 

Labeling segments in an image can be considered as assigning labels to a set of variables, which can be pixels or a group of pixels(super-pixels) while considering (or enforcing) the dependency among the labels.
Several approaches use linear models for this type of problems. Given features of each super-pixel, for instance as an input, the goal is to use inference to find the best assignments of labels to the super-pixels as output. 
Many learning methods have been proposed for structured modeling, which mostly attempt to model dependency among labels during learning process by optimizing a global objective function (such as using Markov Random Field and Conditional Random Field formulations). However, due to efficiency and tractability, these methods are confined to encode only local relationships. 
Although, non-local dependencies can be coded in such a model as well, the model needs to learn more parameters (e.g., in graphical models more edges and weights need to be included to model long-term dependencies). While this can be done by infusing the knowledge in the model as constraints, rather than using a fully connected graph to capture the all possible interactions or higher order representation can be employed, however, in both cases the complexity of method during learning and inference increases. 

\begin{figure*}[t]
\begin{center}
  \centering
  \includegraphics[width=.98\textwidth]{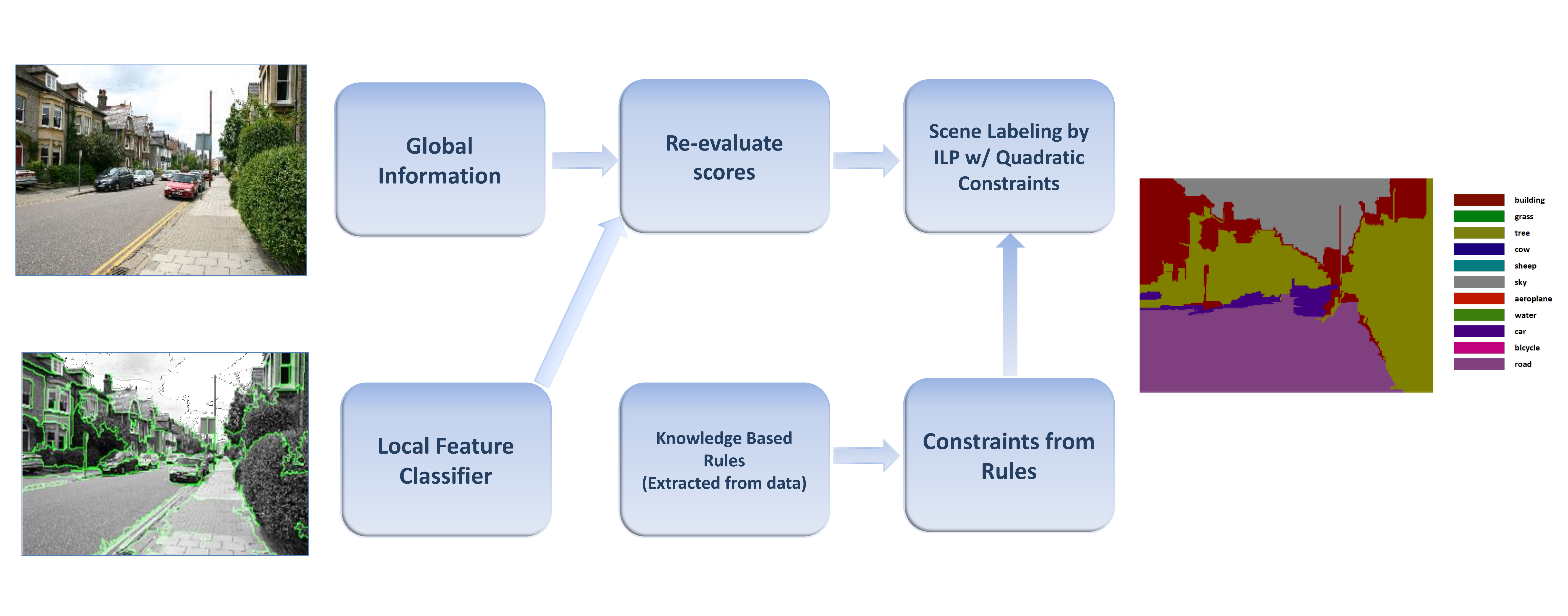}\\
  \caption{   An overview of the proposed approach. For training, we begin by segmenting images into super-pixels and extracting the feature matrix. Then classifiers ( extreme gradient boosting trees) are trained. Also we find the scene-labels association matrix to capture global context. In the inference part during testing, for a given image the label scores are obtained via the classifiers results. Finally, labels scores are updated by applying constraints learned from knowledge-based rules through the optimization function by Integer Programming. }\label{overview}
    \end{center}
\end{figure*}

In this paper, we propose to use the knowledge of some relevant interactions between labels  {\em directly} in the model, instead of {\em indirectly} via learning. In doing so, we benefit from the prior knowledge in inference stage, by adding some constraints that should be satisfied. As a result, we apply inference only in the testing, and we do not require solving any inference problem during the training process. Therefore, we can use any training algorithm, and apply constrained inference model; that way the features and constraints can be distinguished. 
We use a data-driven approach to extract rules to form the training data which is later used to generate expressive constraints. Since the constraints are formed as Boolean functions, they can be represented as logical expressions. By analyzing constraints extracted from the data, we discovered that we need to model soft-constraints in addition to the hard constraints. For instance, most of the time mountain and building are not seen together, but this is not always the case. Thus, we use slack variables in the objective function to model the  soft constraints.   To solve the inference problem with expressive constraints, we use an Integer Programming formulation. %to solve the inference problem. 

 By exploiting declarative constraints, we eliminate the need for re-training the model for new incoming data; because we can simply add more constraints in inference part without changing the prediction problem. Furthermore, we exploit global context of the image by learning scene-label association weights, to increase the probabilities of the most confident labels and limit the number of labels, which need to be explored. 
In order to demonstrate the performance of our method, we report experimental results on three benchmark datasets including SIFTflow \cite{liu2011sift} MSRC2 ~\cite{shotton2009textonboost}  and LMSun~\cite{tighe2013finding}.

In summary, we make following contributions:
\begin{itemize}
\item We use extreme gradient boosting algorithm to learn efficient classifiers to label local features with no need to train detectors or expensive classifiers. 
\item  We improve the scores of super-pixels by combining local classifiers results and global scene information obtained by a learned scene-label association.
\item We incorporate high-level knowledge of relations between labels via imposing rule-based constraints in an Integer Programming problem.
\end{itemize}
The rest of the paper organized as follow: section 2 reviews relative work proposed for scene labeling, in section 3 our proposed method is described in details. The experiments and evaluations of our method are presented in section 4, and finally, we conclude our paper.

\section {Related Work}

Scene labeling has been the interest of many research works recently. Proposed methods differ regarding features and descriptors they use, primitive elements (pixels, patches or regions), employed classifiers and how context techniques are utilized. Conditional Random Fields ~\cite{ladicky2010and} has been used in a vast number of methods. These works use mainly appearance (local features) as unary potential and smoothness between neighboring elements as the pairwise term ~\cite{shotton2009textonboost}. Also using higher order CRF to integrate potential of features at different levels (pixels and superpixels) have been explored ~\cite{russell2009associative}, \cite{gould2009decomposing}. In addition to local features, some methods benefit from object detectors and combine the results from detectors and context information \cite{ladicky2010and}, \cite{tighe2013finding}. 

Other approaches, labels from a dataset of known labels, a retrieval set, transfer to the segments of the query image. To this aim, for a given image, a nearest neighbor algorithm is used to retrieve similar images from a sample data, then a Markov random field (MRF) model is applied on pixels (or super-pixels) in the image to label the pixels ~\cite{liu2011sift} , ~\cite{tighe2010superparsing} and \cite{tighe2014scene}. 
Many works are the extension of this type of labeling, for example, in \cite{eigen2012nonparametric}the weights of descriptors are learned in an off-line manner to reduce the impact of incorrectly retrieved super-pixels. Also, \cite{singh2013nonparametric} proposed to use a locally adaptive distance metric to find the relevance of features for small patches in the image and to transfer the labels from retrieved candidates to small patches of the image. In \cite{gould2012patchmatchgraph} instead of using a retrieval set to transfer the labels, a graph of dense overlapping patch correspondences is constructed; and the query image is labeled by using established patch correspondences. \\

In \cite{isola2013scene} an image is represented as a collage of warped, layered objects obtained from reference images. The scene is analyzed through synthesizing the information. A dictionary of object segment candidates that match the image is retrieved form samples for a test image; then the image is represented by combining these matches. To do so, a dataset of labels exemplars is required. In \cite {guo2014labeling} authors proposed to find the bounding boxes of the objects using object detectors and then regions are classified by combining  information from detectors and surface occlusions; they also use RGB-depth to understand the scene.

In some other works, context information is employed in the modeling of the problem via utilizing global features of the image or incorporating the co-occurrence of the labels in training\cite{vu2013improving}.
Recently, deep learning techniques have been used in scene labeling. For instance, in  \cite{farabet2012scene} for each pixel in the image multi-scale features are gained, and the deep network aggregates feature maps and labels the regions with highest scores.
These models need an extensive data for training; in scene labeling task pixel-wise annotation is expensive work. Also, unlike expressive models [understanding] insight of the relations between labels is hard to achieve \cite{farabet2012scene}, \cite {socher2011parsing}. \\

Most of these approaches neglect to use domain knowledge and the structure -the interactions and dependencies between different parts of the image geometrically (spatially) or conceptually- implied by the images, or they only consider the smoothness of neighboring pixels. While the early researches in computer vision
(e.g., VISIONS \cite{parma1981experiments}) tried to interpret scenes using knowledge from the structure of the image. In this paper, we aim to incorporate modeling constraints into inference to leverage from this type of information. 

Previous works mainly use fully connected graphs to model long-distance dependencies in contrast to local potentials such as MRF models, in which pairwise terms usually are based on the first-order Markov assumption due to reducing the complexity of the inference. However, increasing the number of edges between nodes in a graph causes the inference to be competently expensive. Our goal is to use knowledge-based information to discard the unnecessary edges and make the model simple. We aim to capture the real and common dependencies using the rules extracted from the data instead of letting the CRF finds the weights for these dependencies. Also, our method can be general in a sense that the various types of relations can be formulated in the same framework of rules such as geometrical dependency, non-co-existence, co-occurrence, presence, and adjacency. While in other works (e.g. \cite{ladicky2010graph},  \cite {roy2014scene}. and \cite{rabinovich2007objects} ) only a subset of these constraints are modeled as hard constraints, whereas the rules which extracted from the real data show that some of the constraints are not hard constraints. Therefore, we use soft constraints for those rules that are most of the time valid, but not always. 

%Also, we formulate the problem as integer linear programming with quadratic constraints, and since we can rewrite the constraint as a set of Linear Inequalities.

\section{Approach}
Our system consists of two main phases. The first phase consists of off-the-shelf parts including feature extraction and classifier training based on local features of the sample training images.
The second phase is the inference, in which for a given query image, using scores computed by the classifiers for each possible label, an objective function is maximized such that the constraints are not violated. An overview of our proposed approach is shown in figure~\ref{overview}.

In training, first we segment images using efficient graph-based segmentation \cite{felzenszwalb2004efficient}.
We use this specific method to be consistent with other methods we compare our method with. Next, for each super-pixel, local features, including SIFT, color histogram, mean and standard derivation of color, area and texture, are extracted. Given these local features, classifiers (extreme gradient boosting trees forest) are trained to label super-pixels using their local features. 

%In order to exploit the context of the images,  we also compute GIST ~\cite{oliva2001modeling} features for each image in the training set. \\

In the inference part, we update the classifiers scores by reducing the scores of the assignments which conflict with the defined constraints. We use the product of experts (\cite{hinton1999products}, \cite{chang2012structured}) by combining the probabilities of the label assignments given by the classifiers and the degree of inconsistency provided by the constraints. This can be done by multiplying the probabilities or equivalently adding their logarithm. Note that this is different from using various features and weighting those features. We define a new scoring function:\\
\begin{eqnarray}
 score(\textup{y}^{i},\textup{x}^{j})= \Phi (\textup{y}^{i}|\textup{x}^{j}) \times  {C}   (\textup{y}^{i}) , 
\end{eqnarray}
where $\Phi $ is classifier probabilities and $C$ is constraints violation degrees.\\

We formulate the prediction of a label for each segment in the image by constrained optimization problem, and employ Integer Linear Programing (ILP) with quadratic constraints as an optimization framework. In doing so, we assign a binary label  $y_{i}^{j} \in \{0,1\}$ to a super-pixel $i$,  in order to find the best assignment $\mathcal{Y} = \{y_{1}^{j} , y_{2}^{j} ,..., y_{n}^{j} \}$, where $n$ is the number of super-pixels and $j$ belongs to labels $ {1,...,l}$. Hence, given the classifier score for each super-pixel, we formulate the inference function as 

\begin{eqnarray}
\arg \max_{\textup{y}} \sum_{i,j}\textup{w}(i,j)\ {y}_{i}^{j},
\label{eq:obj_func}
\end{eqnarray}
where $\textup{w}(i,j)$ is the score of assigning label $j$ to super-pixel $i$, provided by our local feature based classifier. This objective function maximize the number of correct predictions for the super-pixels in the image. Toward exploiting context and prior knowledge, we introduce different types of constraints such as non-coexistence, presence and etc, which are explained in \ref{consts} . The first constraint, which should be applied to all instances, enforces that each super pixel gets exactly one label.  
\begin{eqnarray}
\forall i \  \ \sum_{j=1}^{n} {y}_{i}^{j} =1.
\label{eq:oneLabel}
\end{eqnarray}

In following sections, we explain each part of the approach in detail. \\

\subsection{Local Classifiers}
In this section, we explain the first step of our method. In training, we start with segmenting each sample image into super-pixels using efficient graph-based segmentation method \cite{felzenszwalb2004efficient}, followed by computing a feature vector( including, SIFT, color mean) for each super-pixel in the image.
%, so each super pixel would be represented as a feature vector for the classifier with the label of the majority of its pixels. 
We use the same features as used in \cite{tighe2010superparsing}.

We use a sigmoid function to rescale the classifier scores to  give a chance to other classes, beside the one the classifier with a maximum score, to compete during the optimization phase. By doing so, if the classifier mislabels a super-pixel there is a chance that the label may be changed during the inference phase by applying the constraints. 
 We adapt the parameters of the function using the validation sample data.
Also, the sizes (areas) of super pixels are multiplied by the scores to make larger super-pixels more important during the optimization.

%random forest classifiers  \cite{liaw2002classification}
 We use Extreme Gradient Boosting \cite{friedman2001greedy} with softmax objective function to categorize each super-pixel in an image. Since the training data inevitably is noisy,  super-pixels may break the structure of the data,  the bagging using a subset of training examples and subsets of features are used to reduce the effects of the noisy data. %Nevertheless, for the labels for which training sample are limited, SVM may lead to better results. 
Unlike some of the other methods, which train object detectors in addition to the region classifiers, we only use region features and simple classifiers to obtain the initial label scores for super-pixels. 
In our experiments, boosting trees achieved better results in terms of average accuracy among all the classes, even though we discard some of the samples randomly during the training. We discard some samples since  the number of sample data for some classes is enormous.
%, considering each super-pixel as a sample to train the classifier.  

% \usepackage{multirow}

\subsection{Global Context Information }
In performing scene labeling, incorporating scene information, such as the places, can be helpful. For instance, {\em if the probability of being a desert is high for an image presumably the chance of seeing chair or river would be low}. Therefore, we use image level categories to refine the scores of the local classifiers. We use Places CNN model \cite{zhou2014learning} to find the most probable scene semantics in a given image.
Unlike pixel-level annotation, image level annotation is more feasible, thus training a deep network is doable. This deep network is trained on scene labeled images and includes 205 places categories. Using our training data, we learn a mapping between these categories and the label set. To do so, we employ non-negative sparse regression formulation, and  extract a weight matrix $W$  for scene-label association as follow:
\begin{eqnarray}
\min_{W \geq 0} {\left \| Y-WX\right \|_{F}^{2}+\lambda \left \| W\right \|_1},
\label{eq:sparse}
\end{eqnarray}

where $X$ is the confidence score matrix for scene-categories of training images, and $Y$ is a matrix of present labels in corresponding images. In the interest of putting emphasis on learning the mapping between scene categories and smaller super-pixels and rare classes, in $Y$ if  label $l_{i}$ is present we use $1-n(l_{i})/\sum_{j}{n(l_{j})} $  in $i_{th}$ element. The solution of this problem can be efficiently obtained by FISTA algorithm \cite{beck2009fast} which is implemented in SPArse Modeling Software(SPAMS) \cite{mairal2014sparse}.

Then in testing time, for a given image, using scene categories and the weight matrix the most confident labels are obtained and used to refine the label assignment.

\subsection{Extracting Rules and Creating Constraints} \label{consts}
In this section, we describe the types of constraints that we add to the aforementioned optimization problem. Note that even though these constraints can be obtained by common sense knowledge, we explore the training data to discover plausible rules in the context of semantic labeling of images. For instance, the common sense about label \emph{sky} and \emph{building} would be: {\em sky always is above the building}; however, since images are 2D projections of a 3D world this is not always correct. Therefore, we capture these types of constraints as relative constraints with some penalty if they are violated. We call them soft-constraints and describe them in section \ref{soft} \\

The first type of constraints are {\bf geometrical constraints}; whether two labels can have a particular geometrical relationship or not. For example, \emph{sky} always is seen above the \emph{sea}. For each super-pixel, we keep the bounding boxes information, then we apply a rule:  {\em When the label for a super-pixel is sea, its lower super-pixels can not get label sky}. This type of constraints can be formulated as follows:
\begin{eqnarray}
% {y}_{i}^{a}\sum_{k=i+z}^{n}{y}_{i}^{b}  \leqslant 0,
 {y}_{i}^{a}\sum_{j=i+d}^{n}{y}_{j}^{b} \leqslant 0,  
\label{eq:top_const}
\end{eqnarray}
where $a$ and  $b$ respectively are labels of the lower and upper parts of the image,  and $d$ is the displacement of super-pixel $j$ from super-pixel $i$. By exploring the data, we can find hard constraints belonging to this group (e.g (sky, field), (sea, sand), (ceiling, floor)...).

The second type of constraints are {\bf mutually-exclusive}, which represent cases {\em when labels $a$ and $b$,  never occur together in an image}. This can be formulated as follows
\begin{eqnarray}
\sum_{i}y_{i}^{a}\sum_{j}y_{j}^{b}  \leqslant 0.
\label{eq:mute_ex}
\end{eqnarray}

{\bf Presence} is another type of constraints which can be applied. 
%It is noteworthy that, this is different from "co-occurrence", 
According to this constraint, {\em if a certain label $a$ appears in the image, then there should be at least one super-pixel having label $b$}. We can express this constraints as one of the following formulations:

\begin{eqnarray}
\sum_{i}y_{i}^{a}\sum_{i}y_{i}^{b}\geqslant 1, \\
\sum_{i} y_{i}^{a}-\sum_{i,j} y_{i}^{a}y_{j}^{b} \leqslant 0
\label{eq:mute_ex}
\end{eqnarray}

For example, we have found that whenever the label \emph{balcony} appears in the image there exists, at least, one region labeled as a \emph{building}, similarly whenever the label  \emph{pole} appears in the image, at least, one region is labeled as a \emph{road}. Note that; this constraint is different from ``co-occurrence”. In co-occurrence two labels frequently appear together in images; the constraint is not necessarily always valid, thus, it can suitably be formulated as a soft-constraint.\\ 
These rules can be extended, for example, we could add adjacency constraints to enforce the neighboring super-pixels receive the same labels, or even we can apply some constraints on the other features of the specific labels.

We define the rules as described in this part, and then we use a sample data to find relations between classes which follow these rules. We create a cube which with the size of $number of labels \times number of labels \times number of relations$. Each matrix of this cube shows the frequency of a relation happening between classes, and each cell indicates the frequency of a particular relationship between two categories (class labels). For example, to find labels that have a specific relation, we count how many times this relation occurs in a sample data for each pair of classes. If the value is higher than a threshold, we will consider it as a constraint. In addition, we distinguish between soft and hard constraints if the other way around has occurred or not. For instance, about the geometry relation like above, for a pair semantic labels (ceiling, floor) the relations is always true. Therefore, we consider it as hard constraints. While, for (sky, building) there are some images in which the building super-pixels are above sky, thus we consider this relation as soft constraint, it means that the main formulation can tolerate violating this constraint with a penalty.

\subsection{Integer Linear Programming with soft constraints}    \label{soft}

%Not only our rule extraction approach does derive hard constraints, but also it provides soft constraints such as \emph{co-occurrence} and \emph{relative-geometrical} constraints. 

Our proposed approach for extraction of rules not only helps us in deriving the hard constraints, but it also contributes to providing the soft constraints such as \emph{co-occurrence} and \emph{relative geometrical} constraints.

Since some of the rules may not necessarily be satisfied by all the data, we use 0-1 soft-constraint modeling \cite{srikumar2013soft} and define soft constraints by introducing a new binary variable, $z_{k}$, for each constraint and an associated penalty, $c$, which indicates the degree of violation (how the confidence of assignment should reduce when the constraint is not  satisfied). Then objective function in \ref {eq:obj_func} becomes:

\begin{eqnarray}
\arg \max_{y} \sum_{i,j}\textup{w}(i,j)\ {y}_{i}^{j} - \sum_{k}c_{k}(1-z_{k}),
\label{eq:obj_func2}
\end{eqnarray}
which implies that if constraint $k$ is violated $z_{k}$ would be zero and consequently a penalty will be imposed to the optimization function. Moreover, we need to connect the
$z_{k}$  to the constraint $C_{k}$ as $z_{k}\leftrightarrow C_{k}$. We are able to do that by using logical representation and adding these constraints into the objective function. 
For example, \emph{sky mostly is above the building}; however, based on the rules that we have extracted from the database, this is not always true. Therefore, we change the constraint to soft constraint as  \\
\begin{eqnarray}
  \ {y}_{i}^{building}\sum_{j=i+d}^{n}{y}_{j}^{sky} =0 \leftrightarrow z_{k},
\label{eq:obj_bsky}
 \end{eqnarray}
here { $ \leftrightarrow $ is an equivalence for the conditional constraints, meaning that if constraint k is violated, $z_{k}$ will be zero. Consequently, in the equation \ref{eq:obj_func2} $ c_{k}(1-z_{k})$ will be a positive number, which is subtracted from the score. In order to formulate if-else condition in the Integer programming we use the binary variable and add the constraints as inequalities, which is a common practice. }

The penalty values are obtained statistically from the dataset by finding the probability of each case divided by the number of their appearances. For example, {\em sky} and {\em building}  most of the time occur together, but not all the time. Therefore, we find a penalty using the following  formula, which takes into account the frequency of the constraint violations in the data,
\begin{eqnarray}
c_{k}=-\log \frac{P(C_{k}=0)}{P(C_{k}=1)},
\label{eq:penalty}
\end{eqnarray}\\
where $P(C_{k}=0)$ and $P(C_{k}=1)$ are respectively the probability of violating and satisfying the constraint $k$. \\

 \begin{table*}[t]
\begin{center}
\centering
\caption{\textbf{Summery of the rules extracted from the sample data.  }
\label{tab:rules}}
\begin{tabular}{|l|l|l|}
\hline
Name            & Logic representation & examples                                 \\ \hline 

non-coexistence &       $ \mathbf{y}^{l_{1}} \wedge \mathbf{y}^{l_{2}} =0 , \  \mathbf{y}^{k}=(y_{1}^{k}\vee y_{2}^{k} \vee ... \vee y_{n}^{k})$                 & %(awning, sand) or %
(desert, sidewalk)     \\ \hline
   
geometric       &         ${y}_{i}^{l_{1}} \wedge \mathbf{y}_{i+z}^{l_{2}} =0,\  \mathbf{y}^{k}=(y_{i+z}^{k} \vee ... \vee y_{n}^{k}) $ &% (sun above sea)
(road below window)       \\ \hline
    
presence        &             $ {y}_{i}^{l_{1}} \Rightarrow  \mathbf{y}^{l_{2}}       ,\ \mathbf{y}^{k}=(y_{1}^{k}\vee y_{2}^{k} \vee ... \vee y_{n}^{k})$       &  %(pole , road)  
(balcony, building)      \\ \hline
    
co-occurrence    &         $ \mathbf{y}^{l_{1}} \wedge \mathbf{y}^{l_{2}} =1 , \  \mathbf{y}^{k}=(y_{1}^{k}\vee y_{2}^{k} \vee ... \vee y_{n}^{k})$                   & %(sky, sea) , (sky, building)
 (car, road) \\ \hline

adjacency    &    $  {y}_{i}^{l_{1}} \equiv  {y}_{j}^{l_{2}} = \neg ( {y}_{i}^{l_{1}} \oplus {y}_{i}^{l_{2}})=1  $                   & (sky, sun) \\ \hline
% (sky, building) (car, road) 
\end{tabular}
\end{center}
\end{table*}

\begin{figure*}
\begin{center}
  \centering
  \includegraphics[width=1\textwidth]{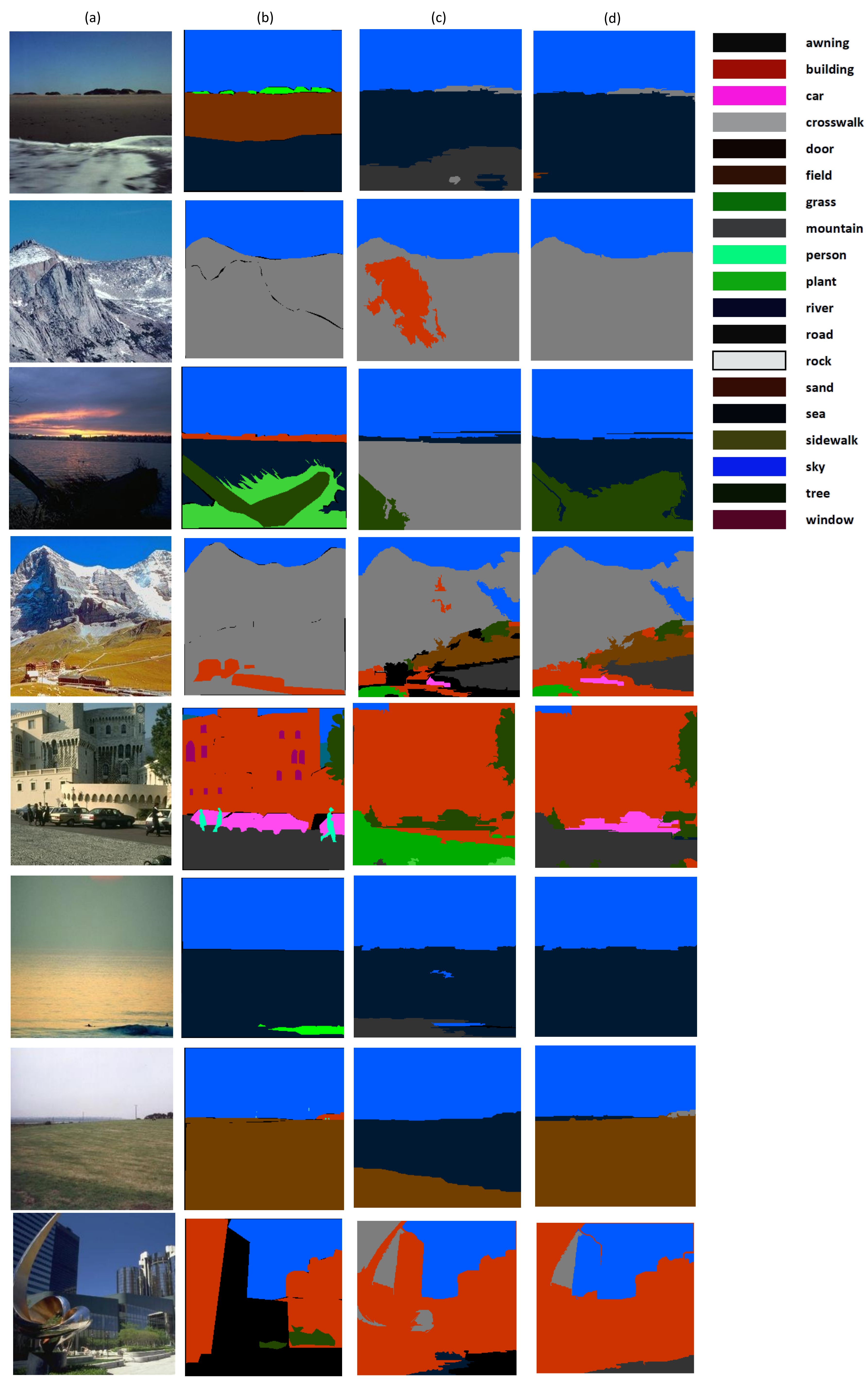}\\
  \caption{{ Examples results obtained by our method on SIFTFlow dataset, (a) query images, (b) ground truths, (c) initial classifier outputs and (d)  
our final results.} }\label{segres}
    \end{center}
\end{figure*}

\subsection{Solving Integer Programming }    
%Even though ILP is in general NP-hard, thanks to many available numerical packages we are able to solve the problem of our size in a short time. 
Thanks to many available numerical packages for estimating Integer Programming solutions, we are able to solve the problem of our size in a short time. We use Gurobi toolkit \cite{gurobi}, which can solve 70 IQP per second on a desktop computer.  The solver uses a piece-wise linear optimization and relaxed LP to solve the integer programming. Also, it is feasible to convert the quadratic constraints to linear ones by adding slack variables due to the boolean nature of our problem. Moreover, one may use other greedy or heuristic methods to solve the objective function.

\begin{figure*}
\begin{center}
  \centering
  \includegraphics[width=1 \textwidth]{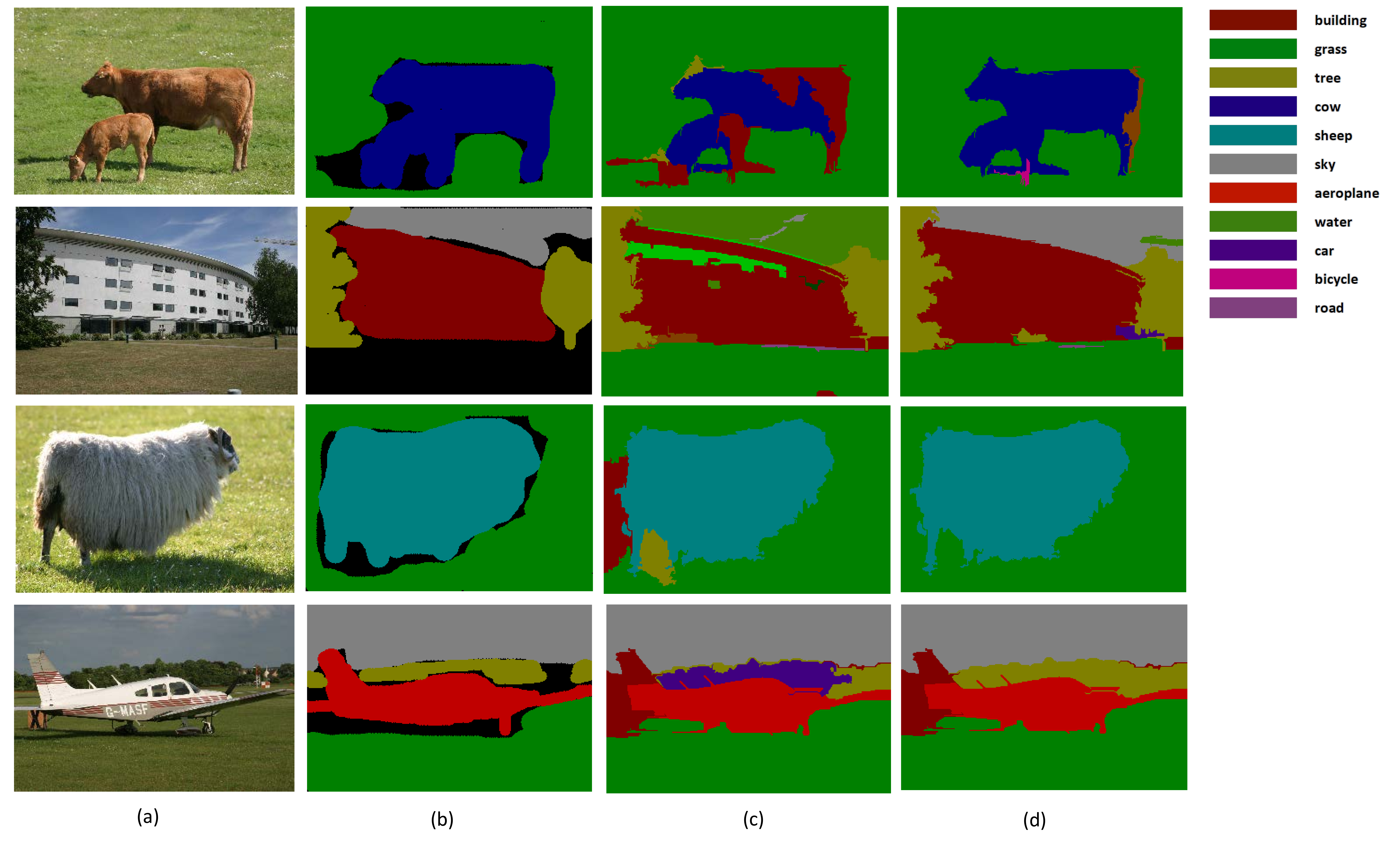}\\
  \caption{{ Example  results obtained by our method on MSRC dataset, (a)   query images, (b)  ground truths, (c)  initial classifier output and (d) our final results.} }\label{msrcfig}
    \end{center}
\end{figure*}

\section {Experiments and Results}
%The first experiment is performed on

We use the SIFTFlow dataset \cite{liu2011sift}, which consists of 2,488 train  and 200 test images collected from LabelMe\cite{russell2008labelme}. These images are from the broad categories such as natural images (coasts, country, etc.) and urban images. Ground truth images are available, in which each pixel of the image is labeled from 33 classes including different objects (e.g., cars, windows) and stuff (e.g., sky, sea).  
In table \ref{tabres}, the accuracy results of our method compared to the state of the similar methods (regarding using similar features) are shown. Our method obtains comparable per-pixel accuracy; even though we do not use object detectors or massively training algorithms.  {Also in table \ref{siftDet2}, we show the accuracy gained in different steps of our method, indicating the improvement due to addition of constraints.} 

In figure \ref{segres}, we show some qualitative results from SIFTflow data set including sample images, ground truth labeling, results from ~\cite{tighe2013finding}, and results obtained by our method before optimization and after applying constraints based optimization. As it is clear, %without using CRF, which requires large training data to determine the weights among the network,
 we are able to improve the results without over smoothing. Also, our method achieves promising results in terms of per-class accuracy as well.
%As it is clear,  in the absence of training a dense CRF to find the weights among the network, we can refine the data without overdoing the smoothness. 
For instance, in the third-row super-pixels below the sea  are labeled as mountain  by our local classifiers, yet constraints such as {\em mountains are not below sea}, are able to handle the miss-labeling and change it to the {\em tree}. Note that plant and tree are very similar in terms of the appearance.
% we will show more results in supplementary materials. \\
\begin{table}[h]
\begin{center}
  \centering
  \caption{{{ Detailed Results on SIFTFlow dataset } }}
\label{siftDet2}
\begin{tabular}{lll}
 \hline \noalign{\smallskip}
{Method }          &  {Avg Accuracy} \\
\hline
\hline
{Local Features + Classifiers } & {72.8}  &  \\
{Local Features + Global Context }  &  {77.3}   &  \\
{Local + Global + Rules }         &   {80.9 } &\\
 \noalign{\smallskip} \hline
\end{tabular}
    \end{center}
\vspace{-5mm}
\end{table}

\begin{table}[h]
\begin{center}
  \centering
  \caption{{Accuracy on SIFTflow dataset  } }
\label{tabres}
\begin{tabular}{llll}
 \hline \noalign{\smallskip}
Method           & Per-Pixel &  Per-Class\\
\hline
\hline
Farabet natural  \cite{farabet2012scene}  & 78.5  & 29.6    &  \\
Farabet balanced  \cite{farabet2012scene} & 74.2  & 46.0   &  \\
Tighe        \cite{tighe2013finding}    & 78.6      &    39.2 & \\
Collage Parsing \cite{tung2014collageparsing} &  77.1   &  41.1   &  \\
Gerorge w/out Fisher Vectors\cite{george2015image}& 77.5 & 47.0 \\
Gerorge Full \cite{george2015image} & 81.7 & 50.1\\
\hline
Ours             &       80.9    &   50.3 &\\
 \noalign{\smallskip} \hline
\end{tabular}
    \end{center}
\vspace{-5mm}
\end{table}
The second data set which we assessed our approach on is LMSun~\cite{tighe2013finding}. This data set contains  45,676
training images and 500 test images including indoor and outdoor scenes, with different image sizes. However, the number of available samples for different labels are not balanced, and some labels are rare. While some labels, for instance, sky and building, have a large number of samples, some others have fewer samples. To make the training more feasible, we use all samples from rare classes and only $25\% $ of samples from common classes for training our extreme gradient boosting classifiers. Also, in learning as well as while using the scene-label associations we assign more weights to smaller super-pixes and rare classes to avoid the influence of common labels such as sky or building.

\begin{table}[h]
\begin{center}
  \centering
  \caption{{Accuracy on LMSun dataset  } }
\label{tablmSun}
\begin{tabular}{l l l l}
 \hline \noalign{\smallskip}
Method           & Per-Pixel &  Per-Class\\
\hline
\hline
Tighe        \cite{tighe2013finding}    & 61.4      &    15.2 & \\
Gerorge w/out Fisher Vectors\cite{george2015image} \ & 58.2 & 13.6 \\
Gerorge Full \cite{george2015image} & 61.2 & 16.0\\
\hline
Ours Local Classifiers & 47.4 & 13.4 \\
Ours Local Classifiers +Global Context & 56.1 & 14.6 \\
Ours             &      58.4   &   17.3 &\\
 \noalign{\smallskip} \hline
\end{tabular}
    \end{center}
\vspace{-5mm}
\end{table}

We also applied our method on MSRCV2 data set \cite{shotton2009textonboost} which has 591 images of 23 classes. We use the provided split, 276 images in training and 255 images for testing. The qualitative results are shown in figure \ref{msrcfig} and our result using the aforementioned approach is presented in table \ref {tabdetailMSRCV}.

\begin{table}[h]
\centering
\caption{Detailed results for MSRCV2 dataset}
\label{tabdetailMSRCV}
\begin{tabular}{llll}
\hline
               Method                                                         & Per Pixel Accuracy \\ \hline \hline
Local Features + Classifiers                                                                 & 76.7              \\ \hline
\begin{tabular}[c]{@{}l@{}}Local Features  + Global Context \end{tabular} & 78.3              \\ \hline
\begin{tabular}[c]{@{}l@{}}Local + Global + Rules\end{tabular}                    & 84.4              \\ \hline
\end{tabular}
\vspace{-4mm}
\end{table}

\section {Discussion}

{The proposed method can get similar results compared to other models such as CRF by having fewer edges i.e. not considering fully connected graph. The two sources of information, including visual features and high-level knowledge-based constraints, are combined to obtain better results. Some relevant information about the data, such as geometrical relationships or non-coexistence, which is hard to learn from the features automatically,  can be easily captured by our proposed method and used to refine the labels. Our approach can scale and generalize. 

In our approach, we add constraints to the trained model, so without retraining, we can add constraints which are declarative and hard to model using only features. Hence, in cases when not enough training data for some categories (class labels) is available, the constraints will help us to obtain better results. Also, when new classes are added to the data set, the proposed method can model the dependencies between classes without learning the pairwise terms since we keep the features (classifier learning) and constraints separate. As shown in experiments on different data sets, with a various number of labels, our method can get promising results. The constraints are assumed to be trusted, and penalties are obtained from data by simply finding the frequency in contrast to learning from the features.} It should be noted that our primary contribution is improving the labeling on top of classifier scores; therefore, using extensive classifiers, such as deep learning, can boost the final results.

\section{Conclusions}
In this paper, we proposed a novel scene labeling approach, in which we use an enhanced inference method that enables the model to incorporate general constraints structure.  We use an integer programming formulation, which can be solved by linear programming relaxation to address the problem. We also proposed to use soft constraints in addition to hard constraints to make the model more flexible.Experimental results on three data sets show the effectiveness of our method.

\section*{References}

%{
%\bibliographystyle{ieee}
\bibliography{Manuscript_Nasim}
%}

%\end{thebibliography}
\end{document}